\documentclass[conference]{IEEEtran}
\IEEEoverridecommandlockouts
\usepackage{cite}
\usepackage{amsmath,amssymb,amsfonts}
\usepackage{algorithmic}
\usepackage{graphicx}
\usepackage{subcaption}
\usepackage{textcomp}
\usepackage{xcolor}
\usepackage{afterpage}
\usepackage{hyperref}
\def\BibTeX{{\rm B\kern-.05em{\sc i\kern-.025em b}\kern-.08em
    T\kern-.1667em\lower.7ex\hbox{E}\kern-.125emX}}
\begin{document}

\title{X-MeshGraphNet: Scalable Multi-Scale Graph Neural Networks for Physics Simulation \\
}

\author{\IEEEauthorblockN{1\textsuperscript{st} Mohammad Amin Nabian}
\IEEEauthorblockA{\textit{NVIDIA} \\
Santa Clara, US \\
mnabian@nvidia.com}
\and
\IEEEauthorblockN{2\textsuperscript{nd} Chang Liu}
\IEEEauthorblockA{\textit{NVIDIA} \\
Santa Clara, US \\
liuc@nvidia.com}
\and
\IEEEauthorblockN{3\textsuperscript{rd} Rishikesh Ranade}
\IEEEauthorblockA{\textit{NVIDIA} \\
Santa Clara, US \\
rranade@nvidia.com}
\and
\IEEEauthorblockN{4\textsuperscript{th} Sanjay Choudhry}
\IEEEauthorblockA{\textit{NVIDIA} \\
Santa Clara, US \\
schoudhry@nvidia.com}
}
\maketitle

\begin{abstract}
Graph Neural Networks (GNNs) have gained significant traction for simulating complex physical systems, with models like MeshGraphNet demonstrating strong performance on unstructured simulation meshes. However, these models face several limitations, including scalability issues, requirement for meshing at inference, and challenges in handling long-range interactions. In this work, we introduce X-MeshGraphNet, a scalable, multi-scale extension of MeshGraphNet designed to address these challenges.

X-MeshGraphNet overcomes the scalability bottleneck by partitioning large graphs and incorporating halo regions that enable seamless message passing across partitions. This, combined with gradient aggregation, ensures that training across partitions is equivalent to processing the entire graph at once. To remove the dependency on simulation meshes, X-MeshGraphNet constructs custom graphs directly from tessellated geometry files (e.g., STLs) by generating point clouds on the surface or volume of the object and connecting k-nearest neighbors. Additionally, our model builds multi-scale graphs by iteratively combining coarse and fine-resolution point clouds, where each level refines the previous, allowing for efficient long-range interactions.

Our experiments demonstrate that X-MeshGraphNet maintains the predictive accuracy of full-graph GNNs while significantly improving scalability and flexibility. This approach eliminates the need for time-consuming mesh generation at inference, offering a practical solution for real-time simulation across a wide range of applications. The code for reproducing the results presented in this paper is available through \href{https://github.com/NVIDIA/modulus/tree/main/examples/cfd/external_aerodynamics/xmeshgraphnet}{NVIDIA Modulus}.
\end{abstract}

\begin{IEEEkeywords}
GNN, simulation, scalability, multi-scale, MeshGraphNet
\end{IEEEkeywords}

\section{Introduction}

\begin{figure}[h]
    \centering
    \begin{subfigure}{0.44\textwidth}  
        \centering
        \includegraphics[width=\textwidth]{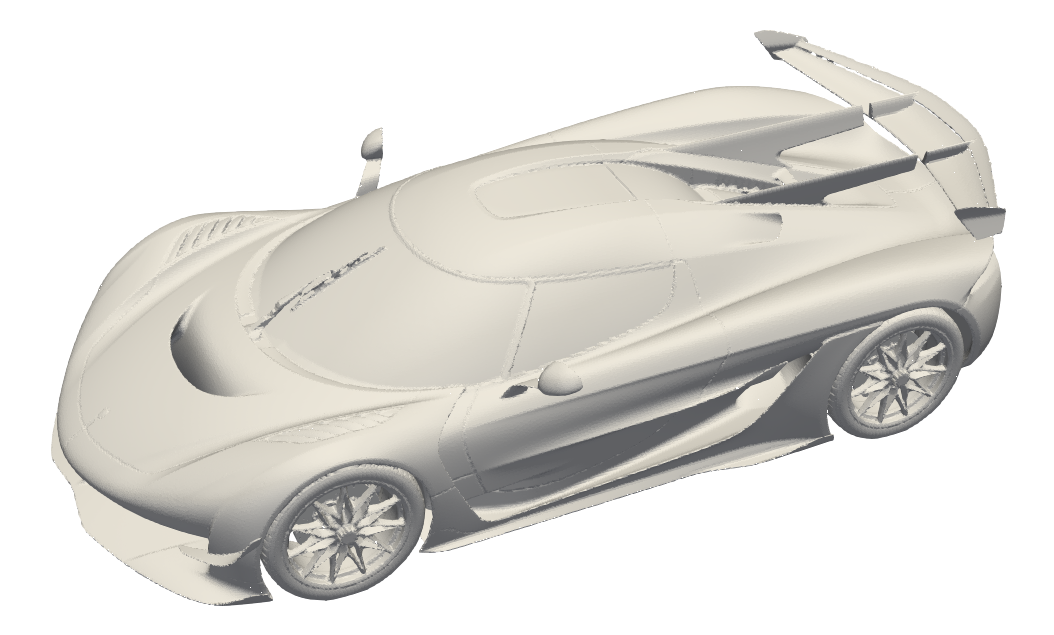}
        \caption{Original tessellated representation.}
        \label{fig:sub1}
    \end{subfigure}
    \begin{subfigure}{0.44\textwidth} 
        \centering
        \includegraphics[width=\textwidth]{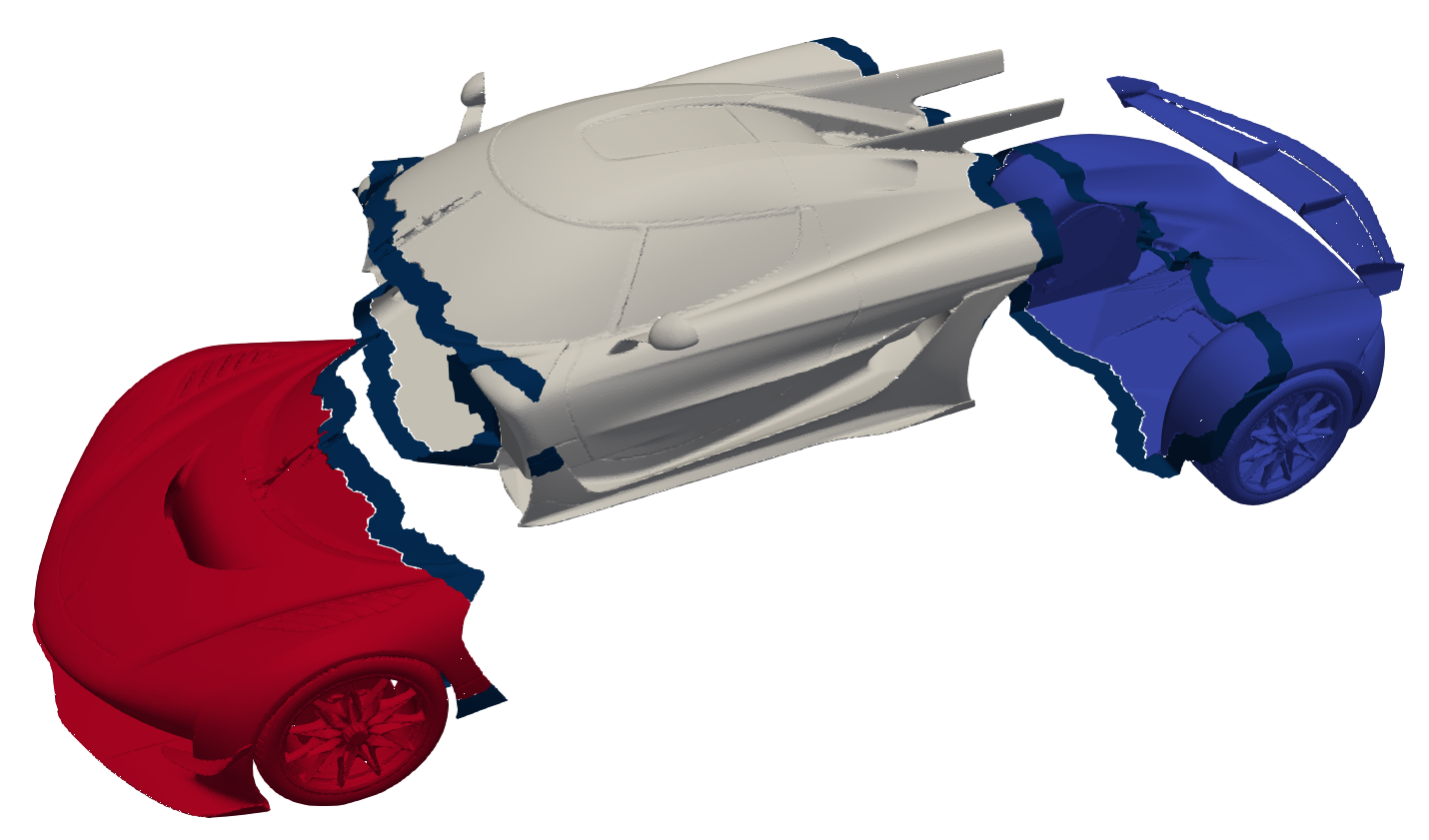}
        \caption{Partitioned tessellated representation with Halo.}
        \label{fig:sub2}
    \end{subfigure}
    \caption{Illustration of the partitioning scheme with Halo on tesselated representation of a Koenigsegg car.}
    \label{fig:partitioning}
\end{figure}

Numerical methods, such as finite element analysis (FEA) and computational fluid dynamics (CFD), offer high-fidelity simulations but come at a significant computational cost. With the growing demand for faster and more efficient simulations—particularly for real-time applications and design optimization—machine learning-based surrogate models have emerged as a promising alternative. Among these, GNNs have garnered attention due to their natural ability to model interactions between components of structured data, such as the meshes commonly used in physical simulations.

\subsection{GNNs for simulation}

In recent years, several GNN models have been proposed as surrogate models for engineering simulations, particularly for systems governed by partial differential equations (PDEs). These models aim to approximate the behavior of complex systems more efficiently than traditional numerical solvers, offering faster inference times while maintaining reasonable accuracy. Early works in this field, such as GraphNet \cite{battaglia2018relational} and Gated Graph Neural Networks (GGNNs) \cite{li2015gated}, have been applied to a wide range of tasks, including fluid dynamics and structural mechanics, where they model physical interactions through message passing between nodes representing simulation elements. More recent models, such as MeshGraphNet \cite{pfaff2020learning} and DeepMind's Learning to Simulate framework \cite{sanchez2020learning}, leverage mesh-based GNNs to represent structured data, where vertices represent simulation nodes and edges represent physical relationships. These models have demonstrated success in tasks such as weather forecasting \cite{lam2022graphcast} and patient-specific cardiovascular simulation \cite{pegolotti2024learning}.

While these GNN-based models show significant promise as surrogates for complex simulations, they still face challenges related to scalability, dependence on simulation meshes, and handling long-range interactions that our proposed X-MeshGraphNet directly addresses.

\textbf{Scalability}: GNN models, including MeshGraphNet, struggle with large-scale simulations due to the high memory overhead associated with processing dense meshes or graphs. As the size of the mesh increases, 
assuming a fixed number of processing layers, both memory and compute scale linearly. If we assume a dynamic number of processing layers (such that the number of layers linearly increases with mesh size, such that the same fraction of the domain is traversed by the GNN kernel), then this scaling is quadratic in both compute and memory.
This limits the applicability of GNN models to high-resolution domains.

\textbf{Dependence on Simulation Meshes}: The reliance on pre-generated simulation meshes during inference presents a significant bottleneck. Generating high-quality meshes for complex geometries is time-consuming and computationally expensive, which undermines the speed advantage of surrogate models.

\textbf{Multi-Scale Simulations}: In multi-scale simulations, different levels of fidelity are required to capture both global and local dynamics. While multi-scale versions of MeshGraphNet have been proposed \cite{fortunato2022multiscale}, they often require handling multiple meshes at various resolutions, which introduces additional complexity and limits usability. Moreover, downsampling and upsampling layers introduce additional memory overhead.

\subsection{Introducing X-MeshGraphNet}

To address these limitations, we propose X-MeshGraphNet, a scalable, multi-scale extension of MeshGraphNet that overcomes the dependency on simulation meshes and improves scalability for large-scale simulations. Our model is based on three key innovations:

\textbf{Graph Partitioning with Gradient Aggregation}: To handle large-scale simulations, X-MeshGraphNet partitions the graph into smaller subgraphs, with halo regions allowing message passing between partitions. Gradients from each partition are aggregated before model updates, ensuring that training remains equivalent to processing the entire graph at once, while maintaining computational efficiency.

\textbf{Custom Graph Construction}: Instead of relying on pre-existing simulation meshes, X-MeshGraphNet generates custom graphs directly from CAD (i.e., STL triangulation) files. This is done by creating a uniform point cloud on the surface or volume of the object and connecting the k-nearest neighbor points, eliminating the need for costly meshing processes during inference.

\textbf{Multi-Scale Graph Generation}: X-MeshGraphNet introduces a multi-scale graph generation process where coarse point clouds are refined iteratively to create finer-scale point clouds, with each level being a superset of the previous. This hierarchical approach allows the model to capture interactions efficiently at various scales.

\subsection{Application to Automotive Aerodynamics}

To demonstrate the effectiveness of X-MeshGraphNet in modeling large-scale simulations, we apply it to an automotive aerodynamics use case. Accurate and efficient prediction of external aerodynamic properties, such as pressure, wall shear stresses, and flow fields, is crucial for vehicle design and optimization. Traditional CFD methods provide high-fidelity results but are computationally expensive, making them impractical for real-time applications and large-scale simulations. Recent advances in deep learning have enabled significant progress in the field of CFD. Several studies have explored the application of deep learning to predict the aerodynamic properties of vehicles. For example, \cite{jacob2021deep} proposes a deep learning model based on a modified UNet architecture \cite{ronneberger2015u} to predict the drag coefficient and velocity field for arbitrary vehicle shapes, using signed distance fields (SDF) as input. This approach offers a significant advantage over traditional methods like Kriging and reduced-order models (ROMs), as it does not require explicit parameterization of the geometry. \cite{chen20213d} proposes another 3D UNet model with a style extractor to obtain deep features of vehicle shapes, with adaptive instance normalization to enhance estimation performance. In addition, a slice-weighted loss function is introduced to improve prediction accuracy, especially in wake regions and near the vehicle surface. This method enables more efficient vehicle shape design by reducing the need for time-intensive CFD simulations. \cite{elrefaie2024drivaernet} proposed the RegDGCNN model, which Leverages the spatial encoding capabilities of PointNet \cite{qi2017pointnet} and the relational inferences provided by graph CNNs. RegDGCNN provides high-precision drag estimates directly from 3D meshes, bypassing traditional limitations such as the need for 2D image rendering or SDF. 

In the realm of generative modeling, \cite{song2023surrogate} integrates physics-based guidance into diffusion models to generate vehicle images with desired aerodynamic properties. By incorporating a surrogate model to estimate drag coefficients, this method allows for the creation of vehicles that are both aesthetically pleasing and aerodynamically efficient. \cite{trinh20243d} introduces a 3D super-resolution model to enhance the resolution of flow field predictions, providing more detailed insight into the aerodynamic behavior of vehicles.

These advances demonstrate the potential of machine learning to overcome the limitations of traditional CFD, offering rapid predictions. However, challenges remain, including the ability to generalize across diverse geometries, handle large-scale simulations, and predict both surface and volumetric aerodynamic quantities.

 In this paper, we leverage X-MeshGraphNet for accurate aerodynamic predictions in passenger car designs. We demonstrate how X-MeshGraphNet scales to large meshes while providing accurate predictions of surface pressure and wall shear stresses for various car designs, effectively addressing key limitations of traditional GNN-based methods, including scalability and dependence on preprocessed meshes. Additionally, we extend X-MeshGraphNet’s underlying halo-based partitioning with gradient aggregation scheme to other neural architectures, including UNets, to highlight the versatility of the proposed approach. To this end, we train a scalable 3D UNet model for volumetric predictions such as pressure and velocity within a bounding box around the car body.

Through experiments conducted on the DrivAerML dataset \cite{ashton2024drivaerml}, these models demonstrate high predictive accuracy and scalability while significantly reducing computational costs compared to standard CFD simulations. The results highlight the potential of X-MeshGraphNet and its extensions as practical and efficient surrogate models for aerodynamic evaluation, supporting real-time applications in automotive design and optimization.

\subsection{Structure of the paper}

In the following sections, we provide an overview of MeshGraphNet and discuss its enhancements through X-MeshGraphNet, focusing on its application to large-scale automotive aerodynamic simulations. We then compare the performance and usability of X-MeshGraphNet with Distributed MeshGraphNet, highlighting key differences in scalability, implementation, and practical deployment. Next, we demonstrate how X-MeshGraphNet’s halo partitioning with gradient aggregation scheme can be extended to other network architectures by training a scalable UNet model with attention gates for volume predictions in automotive aerodynamics.

\section{Technical Overview: MeshGraphNet}

\textit{MeshGraphNet} is a graph-based neural network model designed to simulate physical systems by representing the mesh structure as a graph. The key idea is to use \textit{message passing} between nodes in a graph to propagate information about physical states, such as velocity, pressure, or temperature, over time. The message-passing paradigm enables MeshGraphNet to capture both local and global interactions in a system governed by PDEs.

\subsection{Graph Representation of a Mesh}

In MeshGraphNet, a mesh is treated as a graph $G = (V, E)$, where:
\begin{itemize}
    \item $V$ is the set of nodes, corresponding to the mesh vertices,
    \item $E$ is the set of edges, corresponding to the connections between adjacent vertices.
\end{itemize}

Each node $i \in V$ has a feature vector $h_i \in \mathbb{R}^d$, which stores relevant physical quantities (e.g., velocity, pressure). Each edge $(i, j) \in E$ has a feature vector $e_{ij} \in \mathbb{R}^k$, which encodes information about the relationship between nodes $i$ and $j$, such as distance or relative positions.

\subsection{Message Passing Mechanism}

The central concept of MeshGraphNet is \textit{message passing}, which allows the model to iteratively propagate information between neighboring nodes. The message-passing process involves three main steps: \textit{message computation}, \textit{message aggregation}, and \textit{node update}.

\subsubsection{Step 1: Message Computation}

For each edge $(i, j)$, the model computes a \textit{message} $m_{ij}$ from node $j$ to node $i$. This message is a function of the node features $h_i$ and $h_j$, as well as the edge features $e_{ij}$. The general form of the message function is:

\begin{equation}
    m_{ij} = \phi_m(h_i, h_j, e_{ij}),
\end{equation}
where $\phi_m$ is a learnable neural network that takes the node and edge features as input and outputs a message $m_{ij} \in \mathbb{R}^d$.

\subsubsection{Step 2: Message Aggregation}

After computing the messages from all neighboring nodes $j$ to node $i$, these messages are \textit{aggregated} to update the information at node $i$. The aggregation function can vary, but a common choice is to sum the incoming messages:

\begin{equation}
    m_i = \sum_{j \in \mathcal{N}(i)} m_{ij},
\end{equation}
where $\mathcal{N}(i)$ denotes the set of neighbors of node $i$. The aggregated message $m_i$ contains information from all neighboring nodes.

\subsubsection{Step 3: Node Update}

Once the messages have been aggregated, the node's features are \textit{updated} based on the aggregated message $m_i$. The node update function typically involves a neural network $\phi_u$ that takes the current node features $h_i$ and the aggregated message $m_i$ as input and outputs the updated node features:

\begin{equation}
    h_i' = \phi_u(h_i, m_i),
\end{equation}
where $h_i'$ is the updated feature vector for node $i$. This update allows each node to incorporate information from its neighbors, enabling the model to capture local interactions.

\subsection{Multiple Message-Passing Layers}

The message-passing process described above is repeated over several layers of the graph neural network with distinct network parameters. With each successive layer, information is propagated across progressively larger regions of the graph. After $L$ layers of message passing, each node $i$  aggregates information from nodes up to $L$-hops away in the graph. This allows MeshGraphNet to capture both local and global interactions.

Let $h_i^{(l)}$ denote the node features at layer $l$. The update rule for node $i$ at layer $l$ is:

\begin{equation}
    h_i^{(l+1)} = \phi_u^{(l)}\left( h_i^{(l)}, \sum_{j \in \mathcal{N}(i)} \phi_m^{(l)}\left( h_i^{(l)}, h_j^{(l)}, e_{ij} \right) \right),
\end{equation}
where $\phi_m^{(l)}$ and $\phi_u^{(l)}$ are learnable functions (neural networks) for layer $l$, and $h_i^{(l+1)}$ is the updated node feature vector for node $i$ at layer $l+1$. After $L$-layer message passing, the computational graph for each node's update is self-contained—both forward and backward computations depend solely on its local $L$-hop neighborhood and associated input features. 

\subsection{Application to Physical Simulations}

In physical simulations, the node features $h_i$ typically contain physcial quantities such as velocity $v_i \in \mathbb{R}^3$, pressure $p_i \in \mathbb{R}$, or other relevant physical quantities (e.g., temperature, stress, boundary conditions). The edge features $e_{ij}$ often represent spatial relationships between nodes, such as the relative position vector $x_j - x_i$ or the distance $\| x_j - x_i \|$.

The model is typically trained by minimizing the difference between the predicted physical quantities and the ground truth values obtained from high-fidelity simulations. The loss function typically takes the form of a \textit{mean squared error (MSE)} between the predicted and true values for quantities like velocity, pressure, or displacement.

\section{Methodology}

In this section, we present the methodology underlying X-MeshGraphNet, our scalable multi-scale extension of MeshGraphNet:

\subsection{Graph Partitioning and Gradient Aggregation}
X-MeshGraphNet addresses the scalability challenge of GNN-based simulations by partitioning large graphs into smaller, manageable subgraphs while including sufficient halo region for each partition. This can decouple subgraphs during training, making each subgraph (with additional computations on the halo region) entirely independent. As a result, subgraphs can be treated as separate batches, similar to a Distributed Data Parallel (DDP) setup, which enables the model to handle large-scale simulations that would otherwise be infeasible due to memory constraints.

\textbf{Graph Partitioning:} The input graph, generated from the point cloud, is partitioned into smaller subgraphs. To preserve the information flow between these subgraphs, we introduce a halo region around each partition. The size of the halo region is set to be equal to the number of message passing layers, ensuring that information from neighboring partitions is properly incorporated during training or inference. For efficient graph partitioning, we utilize METIS \cite{karypis1998fast}, a well-known graph partitioning tool. As an example, Figure \ref{fig:partitioning}  shows the original and partitioned mesh graph representation with Halo for a Koenigsegg car STL. By making the number of nodes and edges in each partition similar, we can achieve better load balancing across multiple GPUs, ensuring that each GPU has an equal computational workload, which in turn maximizes resource utilization and improves overall training efficiency.

\textbf{Message Passing Across Partitions}: Each partition performs message passing among its nodes, incorporating input from the halo region to ensure that each node's computational graph matches that of the full graph, thereby guaranteeing mathematical equivalence during forward and backward computations. It is important to note that for this equivalence to hold, computations must remain local. Consequently, operations that rely on global batch statistics, such as batch normalization, are not supported within this framework.

\textbf{Gradient Aggregation}: To ensure that partitioning does not affect the overall training process, we implement a gradient aggregation mechanism. After each training iteration, the gradients from all partitions are aggregated, and the model parameters are updated as if the entire graph had been processed. This technique makes the training process on partitioned graphs equivalent to training on the full graph, while reducing memory overhead.

\subsection{Custom Graph Construction}
To address the dependency on simulation meshes, X-MeshGraphNet generates a custom graph directly from the CAD file of the object being simulated. This is achieved by creating a uniform point cloud on either the surface or volume of the object, based on the simulation requirements. The steps are as follows:

\textbf{Point Cloud Generation}: For each sample, we generate a point cloud by sampling points either on the surface or within the volume. The number of points can be adjusted based on the desired resolution and the complexity of the geometry. Uniform distributions or geometry-aware distributions, such as those based on curvature, can be used for point sampling.

\textbf{K-Nearest Neighbor Connectivity}: Once the point cloud is created, we construct a graph by connecting each point to its k-nearest neighbors. The value of k is chosen to ensure sufficient connectivity for message passing.

This custom graph construction eliminates the need for generating a simulation mesh during inference, which significantly reduces computational overhead and simplifies the pipeline for real-time simulations.

\subsection{Multi-Scale Graph Generation}
To capture dynamics across different radii of influence, X-MeshGraphNet utilizes a multi-scale graph generation process. This involves the creation of point clouds at multiple resolutions, where finer point clouds are iteratively built on top of coarser ones.

\textbf{Point Cloud Generation}: The process begins by generating a point cloud from the CAD file. This representation captures the overall structure of the object, and the graph connectivity is established using k-nearest neighbors, as described earlier.

\textbf{Refining the Point Cloud}: A finer point cloud is generated by increasing the number of sample points. The points from the initial point cloud serve as a subset of the finer point cloud. New connections are then established.

\textbf{Multi-Scale Hierarchical Graphs}: This process is repeated iteratively to produce multiple levels of resolution. At each level, the point cloud from the previous scale is a subset of the point cloud at the next finer scale. Edge connectivity is determined at each scale, ensuring that both local interactions and long-range dependencies are captured across different levels.

\subsection{Model Architecture, Training, and Inference}
X-MeshGraphNet builds upon the architecture of the original MeshGraphNet, using message-passing GNN layers to propagate information across the graph. Standard losses like mean squared error (MSE) are used, but task-specific modifications can be introduced as well, such as physics-informed losses which represent the residual of a specific partial differential equation (PDE). Halo nodes are filtered out before the loss computation. During inference, X-MeshGraphNet follows a similar process to training, except that no ground truth data is provided. The model takes the CAD file of the object, generates the custom graph and the partitions, and predicts the physical quantities of interest. The number of partitions required for inference can be significantly smaller than those used during training, as inference has a much lower memory overhead compared to training. Inference is performed independently on each partition. Predictions on halo nodes are discarded, and the remaining predictions are aggregated on the master rank to reconstruct the full-domain output.

\section{Discussion on Scalability: X-MeshGraphNet vs. Distributed MeshGraphNet}

Distributed MeshGraphNet, an established approach for scaling the MeshGraphNet model to large graphs, uses a distributed message-passing paradigm with all-to-all communication to exchange information among partitions for each message-passing layer \cite{nvidia2023modulus}. While this is often considered a default solution for large-scale simulations, X-MeshGraphNet adopts a more flexible approach by introducing halo regions to handle scalability. This design allows X-MeshGraphNet to distribute computations across multiple GPUs in a simpler and more efficient way, using Distributed Data Parallelism (DDP). Here, we discuss why this approach can be preferable over traditional distributed training strategies.

Usability and Simplicity: In contrast to distributed message passing, which requires intricate setups for device synchronization and communication optimization, X-MeshGraphNet simplifies multi-GPU training by partitioning the graph and using halo regions. This converts the training to a straightforward DDP setup, where each GPU processes its own partition and halo regions ensure sufficient communication between neighboring partitions. The complexity of managing inter-GPU communication and synchronization is greatly reduced compared to other distributed training approaches for GNNs. Furthermore, distributed training frameworks are prone to implementation challenges such as synchronization issues, load balancing, and implementation errors that are difficult to debug. With halo regions, X-MeshGraphNet minimizes these concerns. The use of DDP with halo regions provides a more reliable and easier-to-implement solution, simplifying the overall development and debugging process.

Resource Availability and Flexibility: One of the significant advantages of X-MeshGraphNet’s approach is its adaptability to a wide range of hardware setups. Distributed message passing often requires high-performance interconnects like NVLink or InfiniBand to optimize communication between GPUs. X-MeshGraphNet's DDP setup, augmented with halo regions, reduces the need for such specialized hardware, allowing for efficient training across various interconnect infrastructures, making it more practical and accessible in diverse environments. Moreover, resource availability limits the scale of the model in distributed setups, while partitioning with halo regions imposes no strict requirements. This approach can even enable training on a single GPU, albeit with longer training times due to reduced parallelism.

Optimizing Communication and Scalability: Distributed message passing often requires intensive optimization to minimize communication overhead, particularly as the number of GPUs increases. Ensuring strong and weak scalability under varying network conditions is challenging. In contrast, X-MeshGraphNet's halo regions maintain efficient communication between neighboring partitions and GPUs, simplifying the optimization process. This ensures robust scalability across different network topologies and interconnect infrastructures, without the need for complex communication strategies.

Efficient Multi-GPU Utilization: The halo region method allows X-MeshGraphNet to fully utilize multiple GPUs without introducing the overhead of complex communication protocols. Each GPU processes its local partition, and the halo regions ensure sufficient overlap between partitions for message passing. This strategy leads to better GPU utilization, optimizing both memory usage and computation time without the need for intricate device management.

In summary, while distributed message passing can offer scalability, X-MeshGraphNet's use of halo regions provides a more streamlined and efficient approach, converting the problem into a form of distributed data parallelism. This makes the approach easier to implement, more robust, and adaptable to a broader range of hardware setups. The simplicity and effectiveness of the halo region method make X-MeshGraphNet a practical approach for scaling up MeshGraphNet. In the next section, we include a quantitative analysis comparing the scalability of X-MeshGraphNet with that of Distributed MeshGraphNet.

\section{Case Study: Predicting Aerodynamics on the Surface of Cars using X-MeshGraphNet}
In this section, we demonstrate the application of X-MeshGraphNet to predict key aerodynamic quantities, specifically pressure and wall shear stress, on the surface of cars. The proposed models in the literature often face limitations in terms of scalability and depend on significant mesh downsampling, which can negatively affect prediction accuracy. X-MeshGraphNet provides a promising scalable alternative, capable of handling large-scale aerodynamic simulations without sacrificing accuracy. 

\subsection{Problem Setup}
The goal of this case study is to predict surface pressure and wall shear stress over the body of a car under steady airflow conditions. The car geometry is represented in form of STL triangulation, from which a uniform point cloud is generated on the surface. X-MeshGraphNet constructs a graph from this point cloud, where each point represents a surface element, and the edges are formed by connecting the k-nearest neighbors.

The input features to the model include the 3D positions of surface points, surface normals, and Fourier features \cite{tancik2020fourier}, which are computed as the sine and cosine of the position coordinates with different frequency coefficients. The target quantities for prediction are pressure and wall shear stresses. One can choose to exclude the 3D positions from the input data, and that will likely promote  of the model. 

\subsection{Dataset}
\begin{figure}[htbp]
\centerline{\includegraphics[width=0.5\textwidth]{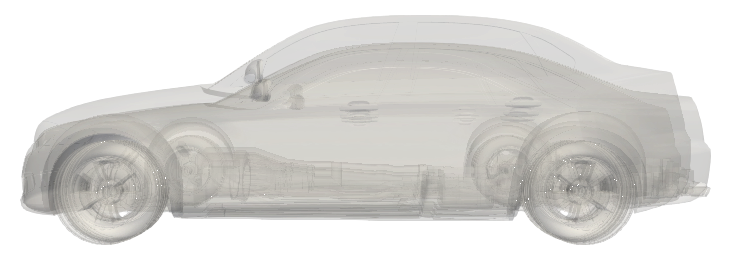}}
\caption{An illustration of the geometry variation in samples 100, 300, and 500.}
\label{dataset_variation}
\end{figure}

For this study, we use the DrivAerML dataset \cite{ashton2024drivaerml}, a publicly available, high-fidelity dataset comprising aerodynamic data for 500 parametrically morphed variants of the DrivAer notchback vehicle. Figure \ref{dataset_variation} shows the geometry variations in three samples. The dataset was generated using hybrid RANS-LES (HRLES), a scale-resolving CFD method, which provides time-averaged quantities for each variant. The available data includes surface pressure, wall shear stress, and flow-field quantities, provided in formats compatible with mesh-based analysis (.vtp for surface data and .vtu for flow-field data). We reserve 10\% of the samples as the test set, with 20\% of the test set consisting of out-of-distribution samples based on drag coefficients. These samples represent extreme cases with the lowest and highest drag coefficients in the entire dataset, which remain unseen by the model during training.

\subsection{Data Preprocessing}
For each car geometry, the .stl files are used to generate a uniform point cloud on the vehicle’s surface. A 3-level graph is constructed with 500k, 1M, and 2M points in the first, second, and third levels, respectively. The corresponding .vtp files are used to read and interpolate pressure and wall shear stress values onto the point cloud using 5-nearest neighbor inverse distance weighted interpolation. Each node is connected to its 6 nearest neighbors. Input and target quantities are normalized using z-score normalization with per-variable global mean and standard deviation. The resulting graphs are divided into 21 partitions with a halo size of 15.

\subsection{Model Training}
The X-MeshGraphNet model is trained using the 24 input features, including Fourier features with 3 different frequencies (i.e., $2 \pi, 4 \pi, 8 \pi$). The Adam optimizer is used with a cosine annealing learning rate schedule ranging from $1 \times 10^{-3}$ to $1 \times 10^{-6}$. The model is trained for 2000 epochs. Gradient clipping with a threshold of $32$ is used. Training is performed in bfloat-16 precision format using Automatic Mixed Precision (AMP). The loss function is defined as the mean squared error between the predicted and ground truth values of pressure and wall shear stress. The model is trained with gradient aggregation applied across graph partitions to ensure consistent model updates, as described in the methodology. Activation checkpointing is also used to reduce the memory overhead. 15 message passing layers are used with a hidden dimension of 512 and SiLU activations.

\subsection{Results}
\begin{figure*}[htbp]
\centerline{\includegraphics[width=0.88\textwidth]{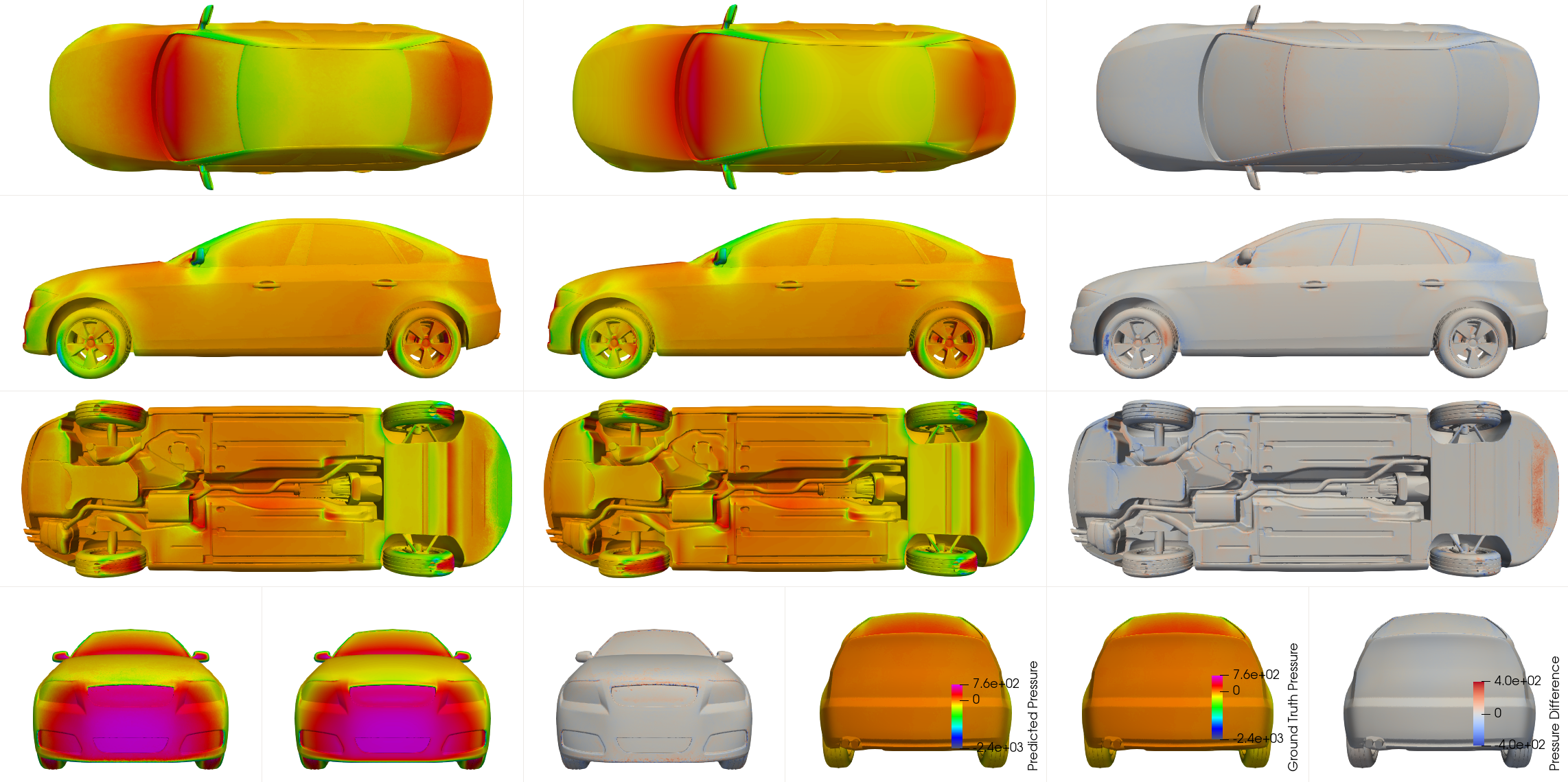}}
\caption{Comparison between the predictions and the ground truth for pressure for Sample 320.}
\label{sample320}
\end{figure*}

\begin{figure*}[htbp]
\centerline{\includegraphics[width=0.88\textwidth]{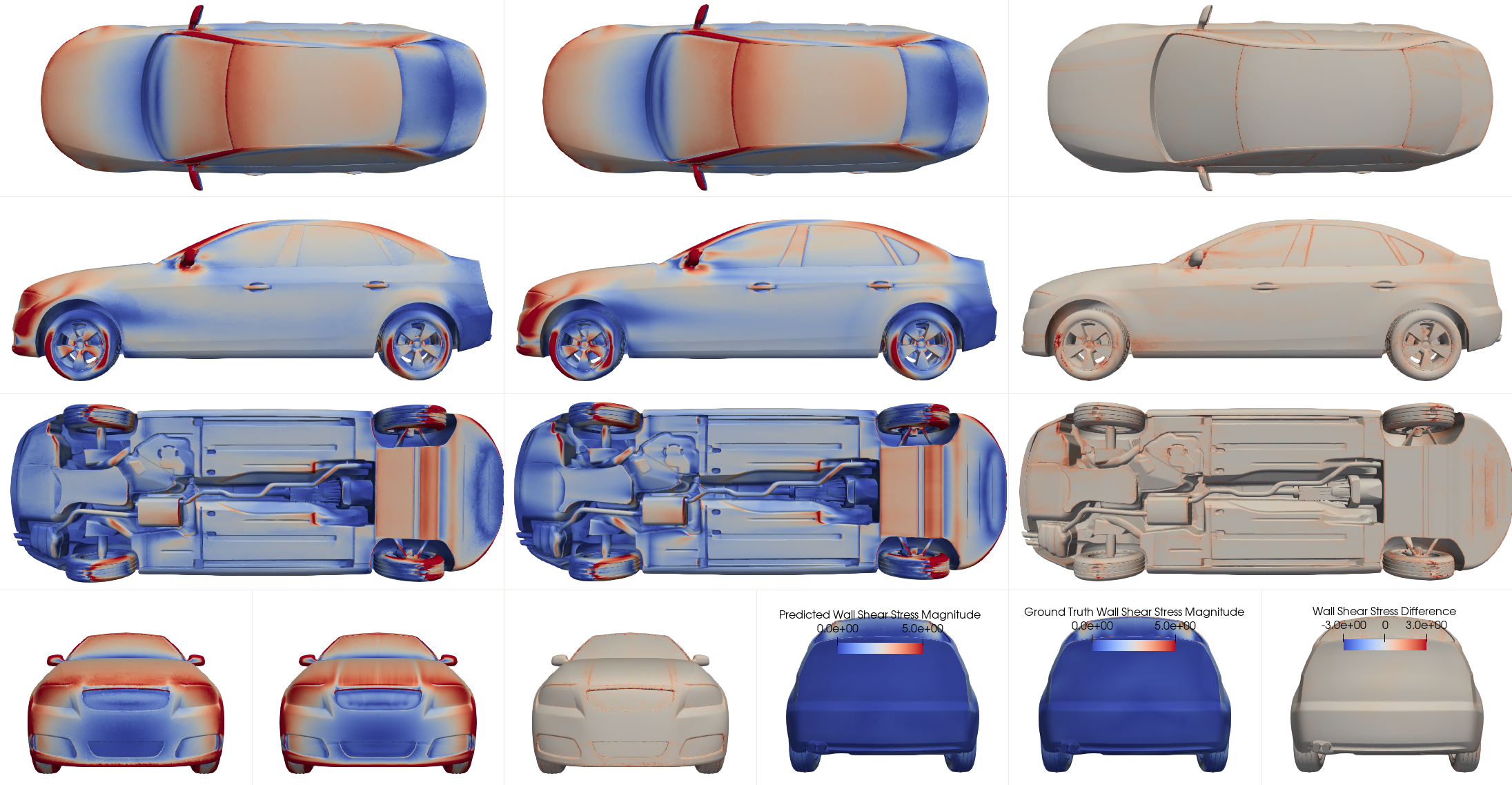}}
\caption{Comparison between the predictions and the ground truth for wall shear stress for Sample 320.}
\label{sample320_wss}
\end{figure*}

\begin{figure}[htbp]
\centerline{\includegraphics[width=0.45\textwidth]{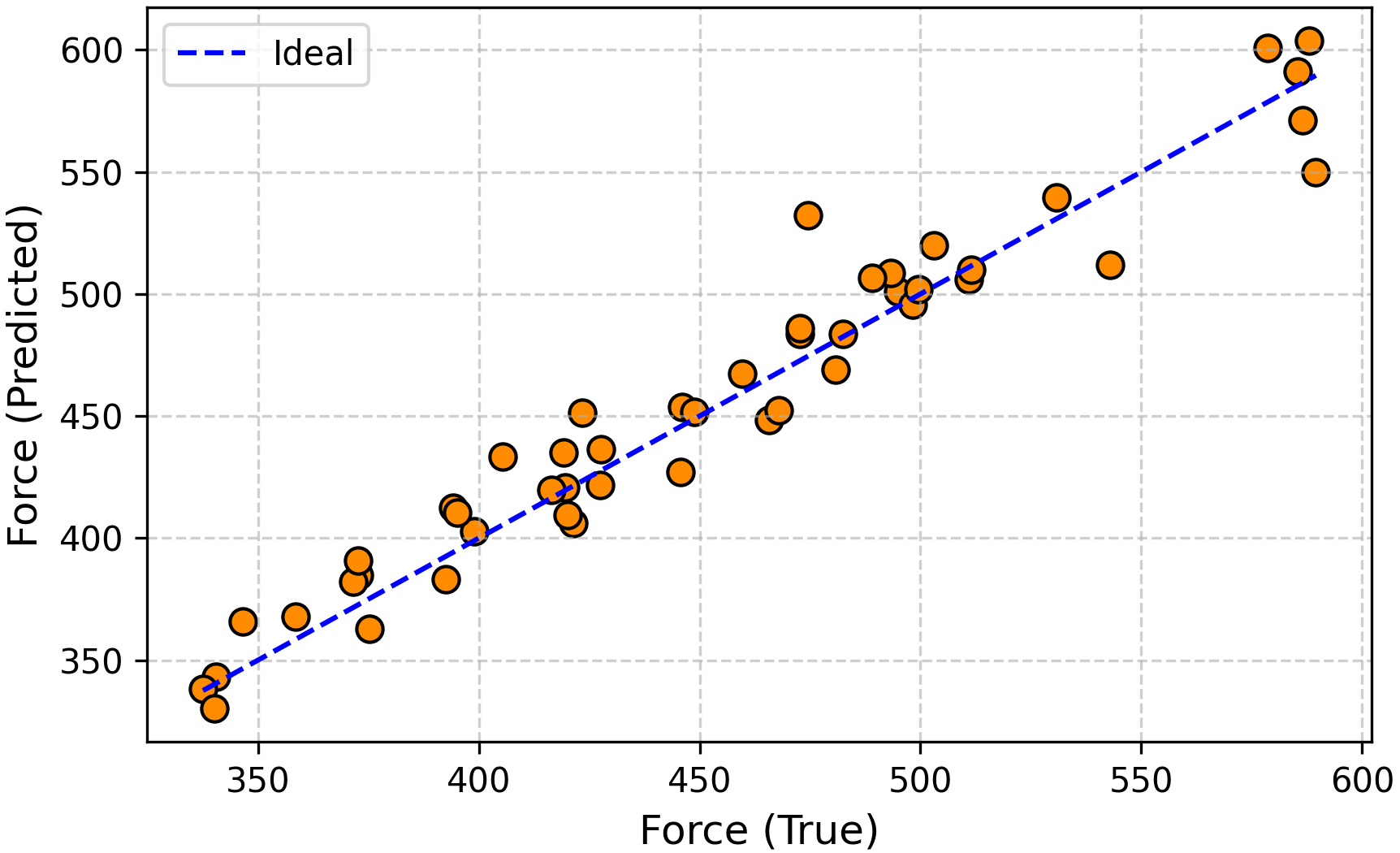}}
\caption{Comparison between the predicted and ground truth force in the flow direction. $R^2$ is $0.942$.}
\label{force_comparison}
\end{figure}

Once trained, X-MeshGraphNet is used to predict the aerodynamic quantities of interest on unseen car geometries. Results for the Sample 320 are shown in Figures \ref{sample320}, \ref{sample320_wss}. The results show that the model is able to capture both the overall aerodynamic profile and localized flow features, such as high-pressure regions at the front of the car and low-pressure areas around the rear. The wall shear stress predictions are also highly correlated with the ground truth CFD simulations, indicating that X-MeshGraphNet can accurately model the tangential forces acting on the car's surface. Moreover, Figure \ref{force_comparison} presents a comparison of predicted versus true aerodynamic forces for different simulation runs in the test set. The dashed blue line represents the ideal case where predictions match the true values perfectly. The coefficient of determination $R^2$ is $0.942$. Table \ref{tab:errors} includes the average relative error for the denormalized predicted quantities.

\begin{table}[ht]
\centering
\caption{Average relative errors for predicted quantities}
\label{tab:errors}
\begin{tabular}{lcc}
\hline
\textbf{Quantity} & \textbf{Relative L2 Error} & \textbf{Relative L1 Error} \\
\hline
Pressure               & 0.1297 & 0.1034 \\
X-Wall Shear Stress    & 0.1678 & 0.1588 \\
Y-Wall Shear Stress    & 0.2199 & 0.2570 \\
Z-Wall Shear Stress    & 0.2774 & 0.2785 \\
\hline
\end{tabular}
\end{table}

\subsection{Performance Analysis}
Memory overhead, particularly activation memory, is a well-known bottleneck in full-graph GNN training methods such as MeshGraphNet, limiting their scalability to larger graphs. X-MeshGraphNet addresses this challenge with a scalable approach that can reduce memory footprint, akin to the well-established data parallelism in distributed training. To evaluate its performance, we used the dataset and model configuration described earlier, testing both single- and multi-GPU setups. In addition to graph partitioning, activation checkpointing with optional CPU-offloading was employed to optimize GPU memory usage, minimizing activation caching by storing checkpoints in host memory.

\subsubsection{Activation Checkpointing with Offloading}
\begin{figure}[htbp]
\centerline{\includegraphics[width=0.5\textwidth]{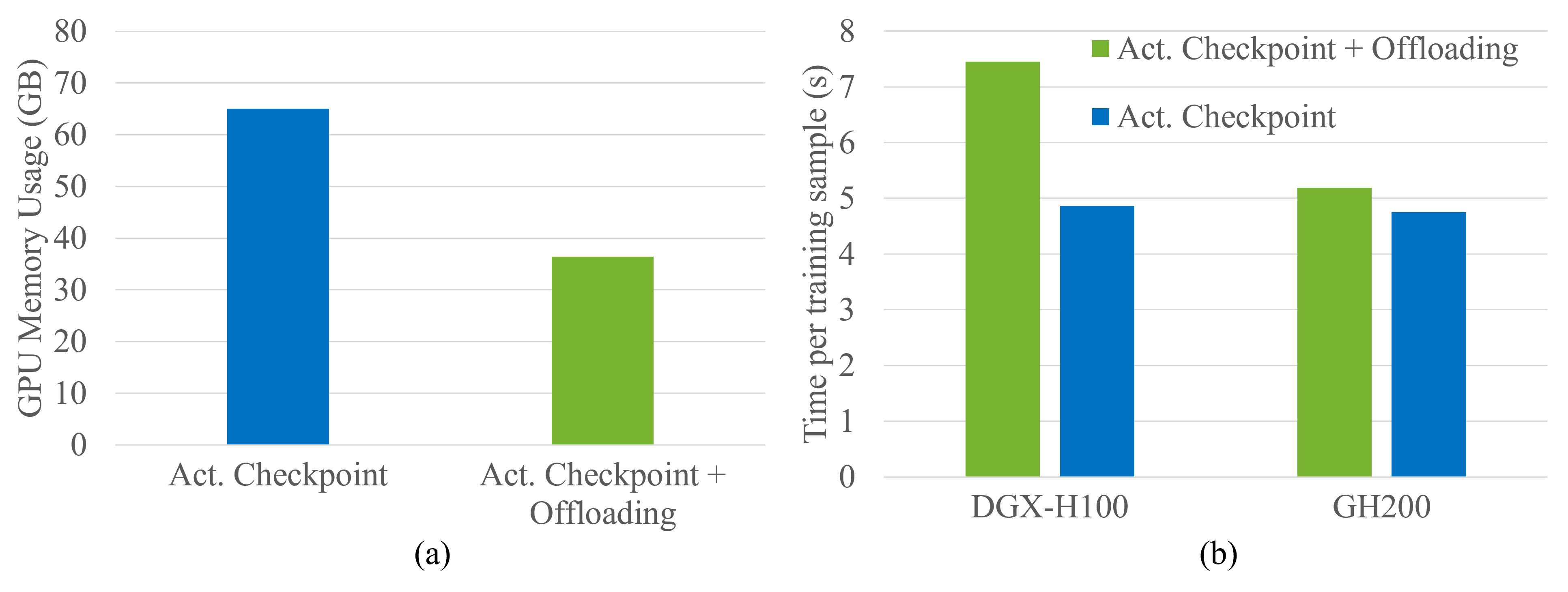}}
\caption{Memory and performance trade-offs between activation checkpointing and activation checkpointing with offloading for a single-GPU run with 2 partitions: (a) GPU memory consumption, and (b) training time (per sample) on DGX-H100 and GH200 systems.}
\label{ac_ao_tradeoff}
\end{figure}

First, we evaluated the trade-offs between activation checkpointing only and activation checkpointing with offloading, on a single GPU with two partitions. As shown in Figure \ref{ac_ao_tradeoff}, offloading activation checkpoints to the CPU reduced memory usage by 1.8×, at the cost of a 1.54× increase in training time on a DGX-H100 machine, and a 1.08× on NVIDIA’s GH200 Grace Hopper Superchip.

\subsubsection{Memory Scaling}
\begin{figure}[htbp]
\centerline{\includegraphics[width=0.5\textwidth]{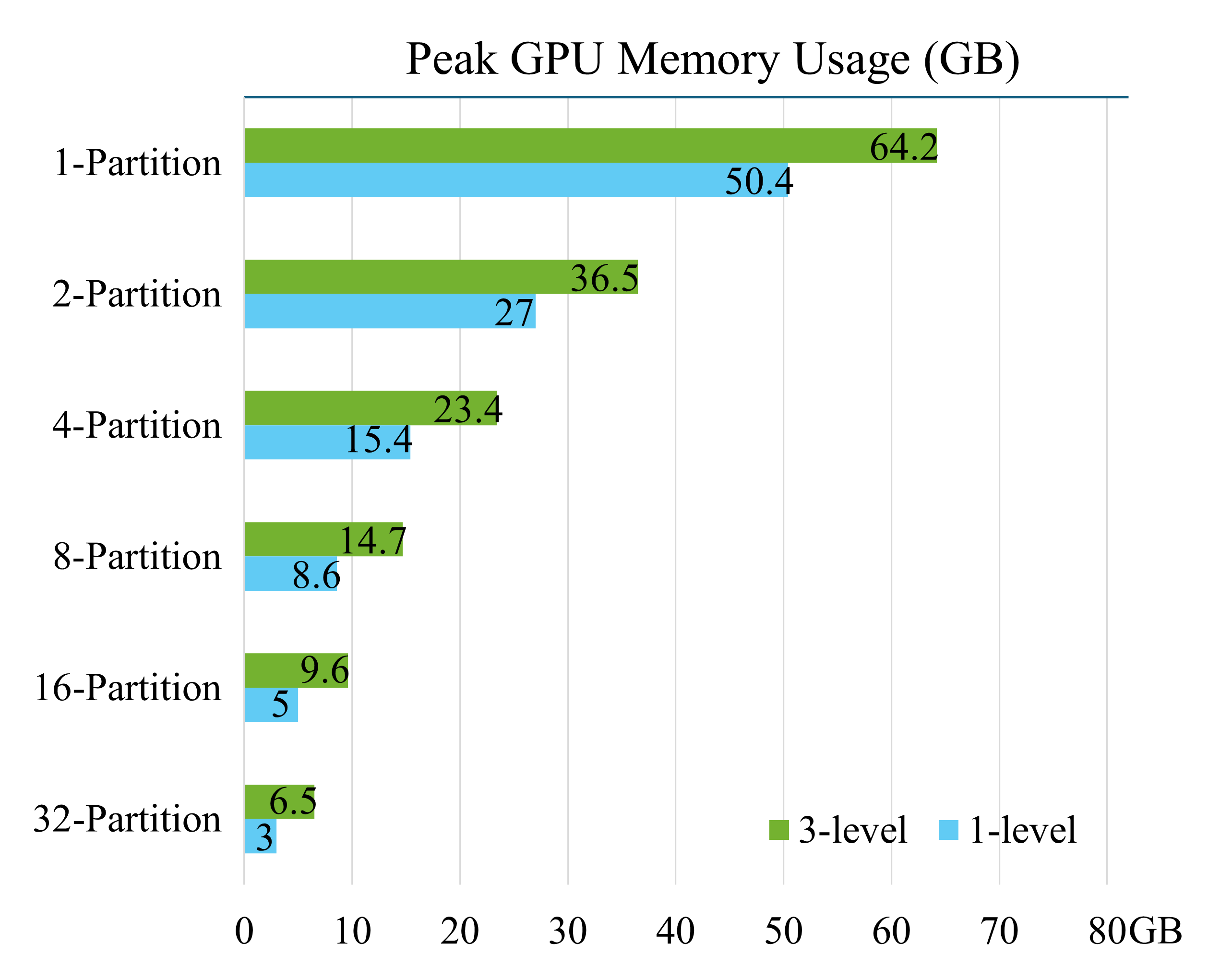}}
\caption{Memory scaling of X-MeshGraphNet on a single-GPU with increasing number of partitions (blue: 1-level graph, green: 3-level graph).}
\label{mem_scaling}
\end{figure}

To assess memory scaling, we increased the number of partitions (including halo regions) to examine whether peak memory usage could be sustainably dropped on a single GPU. Figure \ref{mem_scaling} presents the results using activation checkpointing with offloading \cite{nvidia2023modulus}, which enabled full-graph training (1 partition) to fit within the 80 GB memory limit of a single H100 GPU. As shown in Figure \ref{mem_scaling}, increasing the number of partitions reduced peak device memory usage almost proportionally for both 1-level and 3-level graphs. For example, GPU memory usage dropped from 50.4 GB with 1 partition to 3 GB with 32 partitions on a 1-level graph. While halo regions introduce some memory and computation overhead, they eliminate inter-partition dependencies in full-graph message passing, enabling large graph training on limited GPU resources through smaller partitions.

\subsubsection{Strong Scaling}
\begin{figure}[htbp]
\centerline{\includegraphics[width=0.5\textwidth]{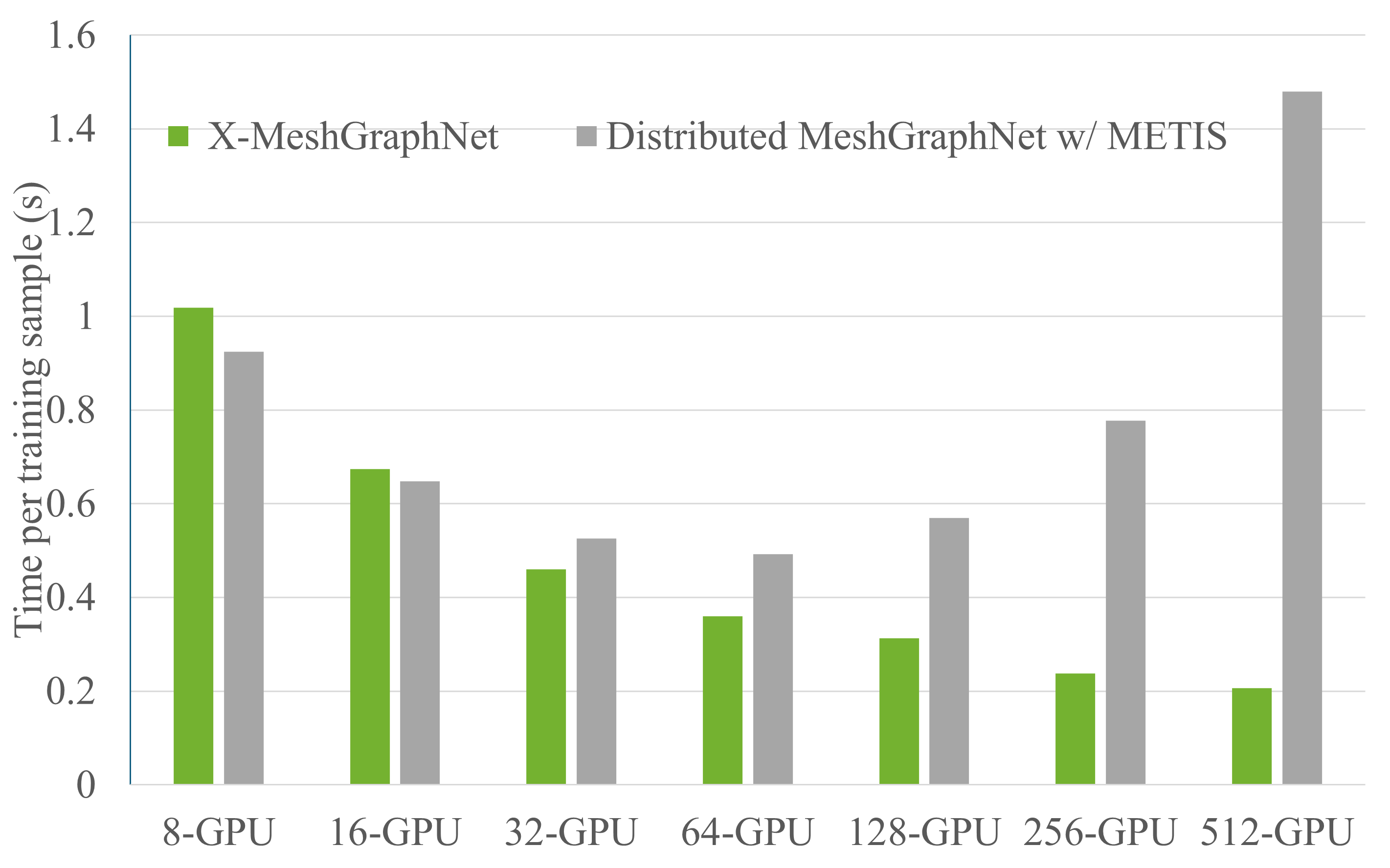}}
\caption{Strong scaling study of X-MeshGraphNet vs. distributed MeshGraphNet \cite{nvidia2023modulus} on the DrivAerML dataset with a 3-level graph of 700,000 nodes. The average training time per sample is measured as the number of ranks (GPUs) scales from 8 to 512.}
\label{scaling_analysis}
\end{figure}

We further performed a strong scaling analysis to compare the performance of X-MeshGraphNet with the distributed message-passing approach (also referred to as distributed MeshGraphNet) from \cite{nvidia2023modulus}.
The distributed message-passing approach in  \cite{nvidia2023modulus} requires all-to-all communication among all GPUs for each message-passing layer across the entire graph, leading to substantial communication overhead at larger scales. The experiments were conducted on an NVIDIA DGX cluster with compute nodes equipped with 8 H100 GPUs (80 GB HBM each) interconnected via NVIDIA Quantum-2 InfiniBand with 400 GB/s bandwidth. A 3-level mesh graph was used with activation checkpointing enabled. For fairness, the number of partitions matched the number of ranks (one GPU per rank), and both methods used the same graph partitioning strategy (METIS). Figure \ref{scaling_analysis} shows that X-MeshGraphNet scales significantly better, consistently reducing runtime as GPU counts scaled up to 512. In contrast, the distributed message-passing approach suffers from increasing communication overhead as the number of GPUs grows, despite efficient graph partitioning to minimize its communication volume.

These findings highlight X-MeshGraphNet’s superior scalability and flexibility, making it well-suited for very large-scale full-graph training on modern GPU clusters.

\subsection{Ablation Study}
\begin{figure*}[htbp]
    \centering
    \begin{subfigure}{0.23\textwidth}  
        \centering
        \includegraphics[width=\textwidth]{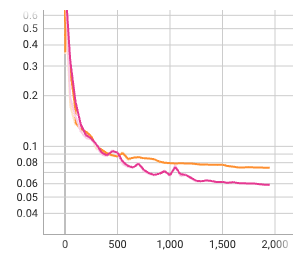}
        \caption{1 vs. 3* levels}
        \label{fig:sub1}
    \end{subfigure}
    \hfill
    \begin{subfigure}{0.23\textwidth} 
        \centering
        \includegraphics[width=\textwidth]{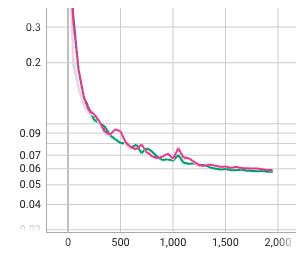}
        \caption{Hidden size of 256 vs. 512*}
        \label{fig:sub2}
    \end{subfigure}
    \hfill
    \begin{subfigure}{0.23\textwidth} 
        \centering
        \includegraphics[width=\textwidth]{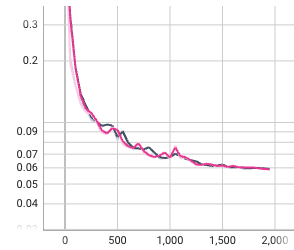}
        \caption{Node degree of 6* vs. 12}
        \label{fig:sub3}
    \end{subfigure}
    \hfill
    \begin{subfigure}{0.23\textwidth} 
        \centering
        \includegraphics[width=\textwidth]{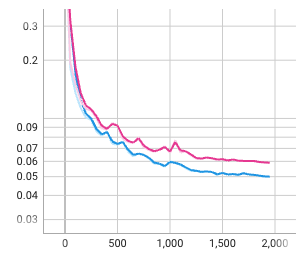}
        \caption{W/* and w/o Fourier features}
        \label{fig:sub4}
    \end{subfigure}
    \caption{Validation loss for different configurations considered in the limited ablation study. Configurations resulting in lower validation error are marked with an asterisk (*).}
    \label{fig:ablation}
\end{figure*}

A limited set of ablation studies was conducted to evaluate the sensitivity of the validation loss to various input variables. These studies included comparisons between a single-level graph and a 3-level graph, hidden sizes of 256 and 512, node degrees of 6 and 12, and training with or without input Fourier features. The results, shown in Figure \ref{fig:ablation}
, indicate that both multi-level training and the inclusion of Fourier frequencies lead to a significant reduction in validation error.

\section{Applicability of the halo partitioning scheme to other architectures}
The halo partitioning with gradient aggregation scheme extends beyond MeshGraphNet and can be effectively applied to other GNNs and also other neural architectures, such as CNNs and UNet-like \cite{ronneberger2015u} models. In these architectures, spatial dependencies between neighboring regions can be captured by treating overlapping regions between partitions as halo regions. This allows seamless information exchange across partitions during both forward and backward passes, ensuring mathematical equivalence to full-domain training. By incorporating halo regions with gradient agregation, large-scale models can be partitioned into smaller, memory-efficient subdomains, enabling scalable training on multi-GPU systems while reducing inter-device communication overhead.

As an example of applying the halo partitioning scheme to architectures beyond GNNs, we introduce a scalable 3D UNet-based model with attention gates, referred to as X-UNet3D in this paper. The model processes volumetric data by leveraging the halo-based partitioning mechanism, enabling large-scale training on multi-GPU systems. X-UNet3D’s architecture features a hierarchical encoder-decoder structure with skip connections and attention gates, allowing it to capture both global context and localized features effectively.

To ensure correct information flow across partitions in X-UNet3D, the halo region size must match the network’s receptive field size, which defines the spatial extent over which a voxel’s output depends on its neighbors. Measuring the receptive field size can be architecture-dependent, and can be done by carefully tracking the data flow during feature propagation through convolutional layers, normalizations, and downsampling or sampling layers. Alternatively, a straightforward empirical approach is to run the network on a full domain and compare its output with that from a partitioned domain using varying halo sizes. The smallest halo size for which the two outputs match indicates the minimum required receptive field size.

Here we use the X-UNet3D model to train a model on the same DrivAerML dataset to predict pressure and velocity in the three-dimensional flow field surrounding the vehicle. Built on a 3-level UNet architecture, our model extends the scalability and efficiency features of the surface model to the volumetric domain, making it a powerful tool for understanding wake dynamics and flow interactions. Unlike the surface model, our volume model operates on a voxelized representation of the volumetric region around the vehicle. A bounding box $[(-3.5, 8.5), (-2.25, 2.25), (-0.32, 3.04)]$ is defined around the vehicle, encompassing the region of interest for aerodynamic analysis. The bounding box is discretized into a uniform voxel grid with a spacing of $1.5 cm$, where each voxel represents a discrete volume element used for predicting aerodynamic properties.

Each voxel's input features include the voxel center coordinates, Fourier features of voxel coordinates (with frequencies of $ \pi, 2 \pi, 4 \pi$), and the SDF along with its spatial derivatives. The model predicts both the velocity and pressure fields. We use Automatic Mixed Precision (AMP) with bfloat16 and apply activation checkpointing to reduce memory overhead. The computational domain is divided into 10 partitions with a halo size of 40. The GeLU activation function is employed throughout the network. The hidden dimension is set to 64 in the first UNet level and is doubled at each subsequent level. The UNet has a depth of 3, with 2 convolutional blocks per level. Each convolutional block uses a kernel size of 3 and a stride of 1, with a pooling size of 2. Attention gates are applied to skip connections. The Adam optimizer is used with a cosine annealing learning rate schedule ranging from $1.5 \times 10^{-4}$ to $5 \times 10^{-7}$. The model is trained for 2000 epochs. In addition to the MSE loss, we also impose a continuity constraint. The required spatial derivatives are computed numerically using the first-order central difference method.

\begin{figure*}[htbp]
\centerline{\includegraphics[width=0.88\textwidth]{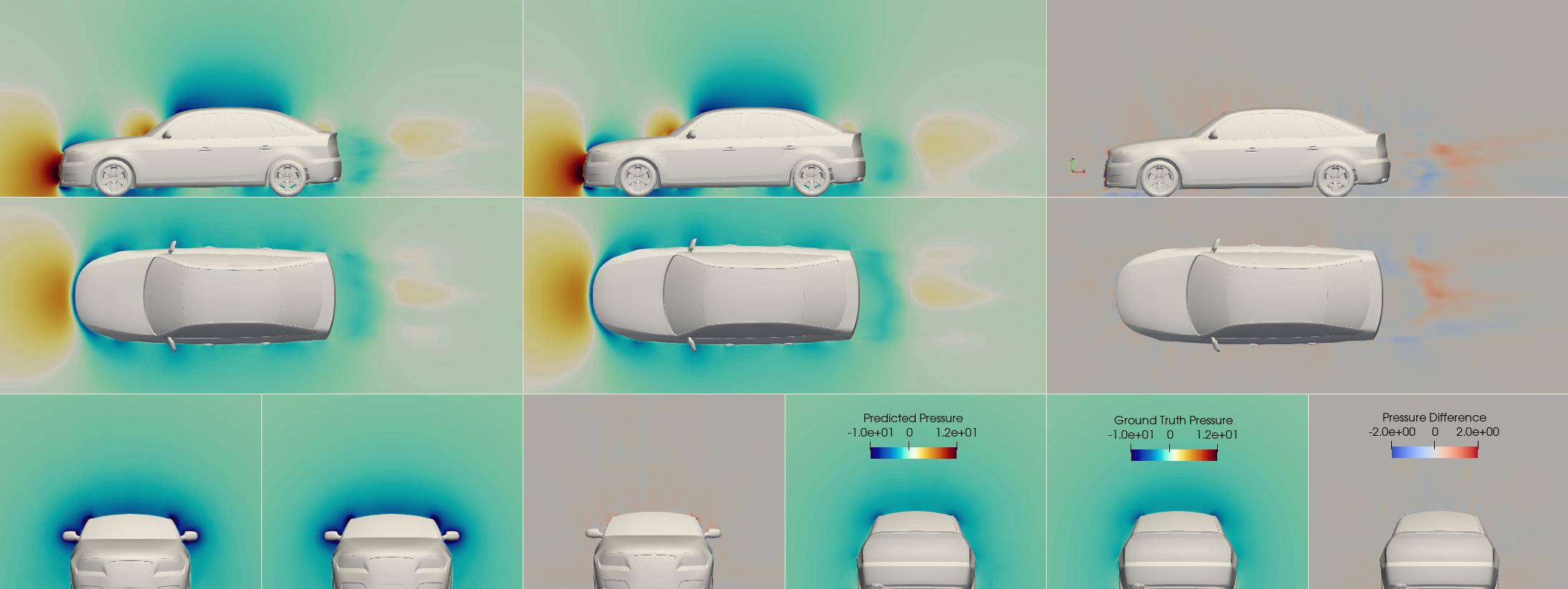}}
\caption{Comparison between the predicted and ground truth pressure for Sample 200.}
\label{sample200_pressure}
\end{figure*}

\begin{figure*}[htbp]
\centerline{\includegraphics[width=0.88\textwidth]{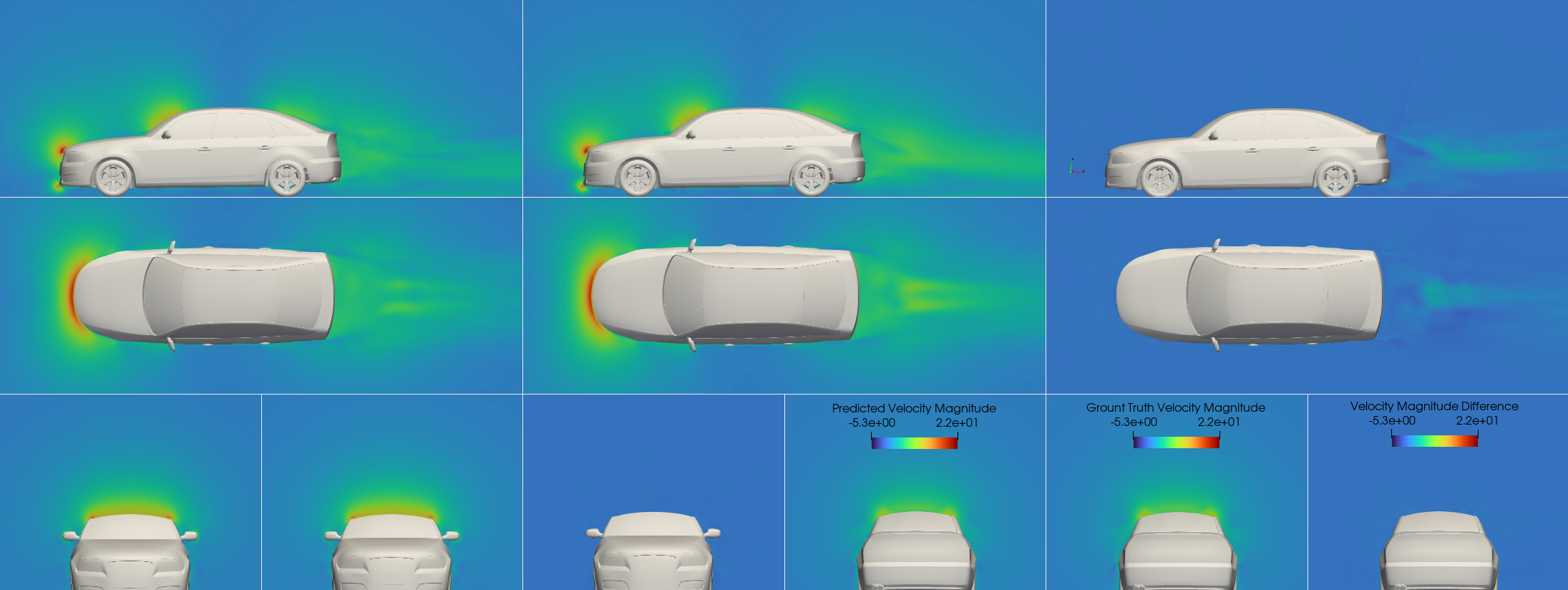}}
\caption{Comparison between the predicted and ground truth velocity magnitude for Sample 200.}
\label{sample200_velocity}
\end{figure*}

Figures \ref{sample200_pressure} and \ref{sample200_velocity} show a comparison between the predicted and ground truth for the pressure and velocity magnitude, respectively. The final MSE loss on the test set is $0.00125$.The model effectively captures wake structures, flow separations, and pressure distributions, providing valuable aerodynamic insights into vehicle performance and stability.

\section{Conclusion}
In this paper, we introduced X-MeshGraphNet, a scalable multi-scale extension of the MeshGraphNet model, designed to address key challenges in GNN-based physical simulations. Our model overcomes the limitations of traditional GNNs in three critical areas: scalability, dependence on simulation meshes, and handling of long-range interactions.

First, X-MeshGraphNet scales to large simulations by partitioning the graph into smaller subgraphs, with halo regions facilitating seamless message passing across partitions. The use of gradient aggregation ensures that the partitioning does not compromise the accuracy or effectiveness of the training process, making it equivalent to training on the full graph while significantly reducing memory and computational overhead.

Second, to eliminate the time-consuming reliance on pre-generated simulation meshes, X-MeshGraphNet constructs custom graphs directly from CAD files. By generating uniform point clouds and connecting k-nearest neighbors, the model avoids the need for meshing, allowing for more efficient and flexible inference.

Third, the model introduces a multi-scale graph generation approach that captures both global and local dynamics by iteratively combining point clouds at different levels of resolution. This hierarchical representation ensures that X-MeshGraphNet can handle the complexities of multi-scale phenomena while remaining computationally efficient.

Our experiments demonstrate that X-MeshGraphNet can maintain the predictive accuracy of traditional GNN models while significantly improving scalability and flexibility. The model’s ability to handle large, complex simulations without requiring meshes at inference makes it particularly well-suited for real-time applications across various domains, from fluid dynamics to structural mechanics.

In future work, we aim to explore further optimizations in graph partitioning and edge connectivity strategies, as well as extend X-MeshGraphNet to handle a broader range of physical simulations, including those involving dynamic or deformable geometries. Additionally, integrating advanced physical constraints into the learning process may further improve the model’s accuracy and applicability to real-world systems.

Future work will also focus on exploring several extensions to further enhance the model's flexibility and performance. One area of investigation will be comparing the effects of constructing graphs using the K-NN approach versus connecting points within a specified radius. Additionally, we plan to experiment with graph augmentation techniques, such as dynamically sampling point clouds and constructing the graph on the fly per epoch. This approach could help mitigate topological biases that arise from fixed graph structures.

Moreover, we intend to explore generating the point cloud non-uniformly, taking into account the curvature information of the geometry. By increasing point density in regions of high curvature, we expect to capture finer details more effectively, potentially improving the accuracy of the model in regions with complex phenomena.

Lastly, we plan to push the limits of scalability by training models with increasing numbers of graph nodes and edges. This will help us understand how far the scalability of X-MeshGraphNet can be practically extended in terms of handling larger graphs.

\section*{Acknowledgment}
The authors would like to thank Peter Sharpe for his valuable comments on this work, and Kaustubh Tangsali for his assistance with the analysis of results for the automotive aerodynamic models.

\bibliographystyle{ieeetr} 
\bibliography{xmeshgraphnet.bib}

\begin{thebibliography}{10}

\bibitem{battaglia2018relational}
P.~W. Battaglia, J.~B. Hamrick, V.~Bapst, A.~Sanchez-Gonzalez, V.~Zambaldi, M.~Malinowski, A.~Tacchetti, D.~Raposo, A.~Santoro, R.~Faulkner, {\em et~al.}, ``Relational inductive biases, deep learning, and graph networks,'' {\em arXiv preprint arXiv:1806.01261}, 2018.

\bibitem{li2015gated}
Y.~Li, D.~Tarlow, M.~Brockschmidt, and R.~Zemel, ``Gated graph sequence neural networks,'' {\em arXiv preprint arXiv:1511.05493}, 2015.

\bibitem{pfaff2020learning}
T.~Pfaff, M.~Fortunato, A.~Sanchez-Gonzalez, and P.~W. Battaglia, ``Learning mesh-based simulation with graph networks,'' {\em arXiv preprint arXiv:2010.03409}, 2020.

\bibitem{sanchez2020learning}
A.~Sanchez-Gonzalez, J.~Godwin, T.~Pfaff, R.~Ying, J.~Leskovec, and P.~Battaglia, ``Learning to simulate complex physics with graph networks,'' in {\em International conference on machine learning}, pp.~8459--8468, PMLR, 2020.

\bibitem{lam2022graphcast}
R.~Lam, A.~Sanchez-Gonzalez, M.~Willson, P.~Wirnsberger, M.~Fortunato, F.~Alet, S.~Ravuri, T.~Ewalds, Z.~Eaton-Rosen, W.~Hu, {\em et~al.}, ``Graphcast: Learning skillful medium-range global weather forecasting,'' {\em arXiv preprint arXiv:2212.12794}, 2022.

\bibitem{pegolotti2024learning}
L.~Pegolotti, M.~R. Pfaller, N.~L. Rubio, K.~Ding, R.~B. Brufau, E.~Darve, and A.~L. Marsden, ``Learning reduced-order models for cardiovascular simulations with graph neural networks,'' {\em Computers in Biology and Medicine}, vol.~168, p.~107676, 2024.

\bibitem{fortunato2022multiscale}
M.~Fortunato, T.~Pfaff, P.~Wirnsberger, A.~Pritzel, and P.~Battaglia, ``Multiscale meshgraphnets,'' {\em arXiv preprint arXiv:2210.00612}, 2022.

\bibitem{jacob2021deep}
S.~J. Jacob, M.~Mrosek, C.~Othmer, and H.~K{\"o}stler, ``Deep learning for real-time aerodynamic evaluations of arbitrary vehicle shapes,'' {\em arXiv preprint arXiv:2108.05798}, 2021.

\bibitem{ronneberger2015u}
O.~Ronneberger, P.~Fischer, and T.~Brox, ``U-net: Convolutional networks for biomedical image segmentation,'' in {\em Medical image computing and computer-assisted intervention--MICCAI 2015: 18th international conference, Munich, Germany, October 5-9, 2015, proceedings, part III 18}, pp.~234--241, Springer, 2015.

\bibitem{chen20213d}
F.~Chen and K.~Akasaka, ``3d flow field estimation around a vehicle using convolutional neural networks.,'' in {\em BMVC}, p.~396, 2021.

\bibitem{elrefaie2024drivaernet}
M.~Elrefaie, F.~Ahmed, and A.~Dai, ``Drivaernet: A parametric car dataset for data-driven aerodynamic design and graph-based drag prediction,'' in {\em International Design Engineering Technical Conferences and Computers and Information in Engineering Conference}, vol.~88360, p.~V03AT03A019, American Society of Mechanical Engineers, 2024.

\bibitem{qi2017pointnet}
C.~R. Qi, H.~Su, K.~Mo, and L.~J. Guibas, ``Pointnet: Deep learning on point sets for 3d classification and segmentation,'' in {\em Proceedings of the IEEE conference on computer vision and pattern recognition}, pp.~652--660, 2017.

\bibitem{song2023surrogate}
B.~Song, C.~Yuan, F.~Permenter, N.~Arechiga, and F.~Ahmed, ``Surrogate modeling of car drag coefficient with depth and normal renderings,'' in {\em International Design Engineering Technical Conferences and Computers and Information in Engineering Conference}, vol.~87301, p.~V03AT03A029, American Society of Mechanical Engineers, 2023.

\bibitem{trinh20243d}
T.~L. Trinh, F.~Chen, T.~Nanri, and K.~Akasaka, ``3d super-resolution model for vehicle flow field enrichment,'' in {\em Proceedings of the IEEE/CVF Winter Conference on Applications of Computer Vision}, pp.~5826--5835, 2024.

\bibitem{ashton2024drivaerml}
N.~Ashton, C.~Mockett, M.~Fuchs, L.~Fliessbach, H.~Hetmann, T.~Knacke, N.~Schonwald, V.~Skaperdas, G.~Fotiadis, A.~Walle, {\em et~al.}, ``Drivaerml: High-fidelity computational fluid dynamics dataset for road-car external aerodynamics,'' {\em arXiv preprint arXiv:2408.11969}, 2024.

\bibitem{karypis1998fast}
G.~Karypis and V.~Kumar, ``A fast and high quality multilevel scheme for partitioning irregular graphs,'' {\em SIAM Journal on scientific Computing}, vol.~20, no.~1, pp.~359--392, 1998.

\bibitem{nvidia2023modulus}
{Modulus Contributors}, ``{NVIDIA Modulus: An open-source framework for physics-based deep learning in science and engineering},'' 2023.
\newblock [Online]. Available: {\url{https://github.com/NVIDIA/modulus}}.

\bibitem{tancik2020fourier}
M.~Tancik, P.~Srinivasan, B.~Mildenhall, S.~Fridovich-Keil, N.~Raghavan, U.~Singhal, R.~Ramamoorthi, J.~Barron, and R.~Ng, ``Fourier features let networks learn high frequency functions in low dimensional domains,'' {\em Advances in neural information processing systems}, vol.~33, pp.~7537--7547, 2020.

\end{thebibliography}

\end{document}